# SLIC Based Digital Image Enlargement

M.Z.F.Amara, R.Bandara, Thushari Silva

*Faculty of Information Technology,*
*University of Moratuwa,*
*Sri Lanka*

amarazuffer@gmail.com
ravimalb@uom.lk
thusharip@uom.lk

# SLIC Based Digital Image Enlargement

M.Z.F.Amara, R.Bandara, Thushari Silva

*Faculty of Information Technology,*

*University of Moratuwa,*

*Sri Lanka*

`amarazuffer@gmail.com`

`ravimalb@uom.lk`

`thusharip@uom.lk`

*Abstract*— Low resolution image enhancement is a classical computer vision problem. Selecting the best method to reconstruct an image to a higher resolution with the limited data available in the low-resolution image is quite a challenge. A major drawback from the existing enlargement techniques is the introduction of color bleeding while interpolating pixels over the edges that separate distinct colors in an image. The color bleeding causes to accentuate the edges with new colors as a result of blending multiple colors over adjacent regions. This paper proposes a novel approach to mitigate the color bleeding by segmenting the homogeneous color regions of the image using Simple Linear Iterative Clustering (SLIC) and applying a higher order interpolation technique separately on the isolated segments. The interpolation at the boundaries of each of the isolated segments is handled by using a morphological operation. The approach is evaluated by comparing against several frequently used image enlargement methods such as bilinear and bicubic interpolation by means of Peak Signal-to-Noise-Ratio (PSNR) value. The results obtained exhibit that the proposed method outperforms the baseline methods by means of PSNR and also mitigates the color bleeding at the edges which improves the overall appearance.

## I. INTRODUCTION

Images are the most convenient form to convey information to humans than plain text. Thus, the visual quality of images needs to be at a considerable level since the information that it holds is capable of delivering meaningful insights. The demand for quality images has risen in the recent past due to its application in a variety of fields including astronomy, satellite imagery, medical imaging, security, computer vision problems and deep learning applications etc. [1-4]. Low resolution (LR) images can be a result from low quality image capturing devices and output from specific image processing applications etc. which may have to be enlarged for visualization purposes or further analysis.

The conversion from LR to high-resolution (HR) image can be done in several ways namely, fusing the information from multiple LR images to reconstruct the details in HR, learning the image features by observing sample pairs of LR and HR images to be used in reconstructing HR image from a single LR image, and as the third method, reconstruct the HR image by interpolating the pixels in LR images [5-7]. There are only limited occasions which encounter a LR image with multiple shots hence the reconstruction of HR image out of multiple LR is not a practical and universal approach. Although the details of natural images can be learnt and then used in enlarging a single LR image, the approach is limited due to the domain dependency and the requirement of excessive amount of training time and training data [8-10]. The pixel interpolation is still the most frequently used technique in many applications. Therefore, in this paper we address some problems encountered in interpolation-based image enlargement for improving the resultant images. The interpolation-based image enlargement techniques generally use only a single LR source image. Therefore, there is a constrain of the amount of information that can be obtained to be used in the reconstruction phase. The procedure followed to find the missing pixel intensity values of the HR image decides the overall quality of the final reconstructed image.

Digital image enlargement involves in digital signal processing that uses different algorithms to enrich different features of an image. Image interpolation, super resolution [11], wavelet-based image interpolation [12] are some examples. Traditional interpolation techniques approximate the values for new points using certain mathematical functions. Nearest-neighbor, bilinear, bicubic and B-splines etc. are a few techniques among them [13]. These techniques can be directly applied to images in any color model. While nearest-neighbor duplicates the same intensity value across nearest pixels, bilinear and bicubic produce new shades (intensities) by using color information of pixels residing in near vicinity of the target pixel. This drawback occurs in the enlargement process introducing the color bleeding effect which is addressed through this work.

Color bleeding takes place when colors leak over the strong edges that separate contrasting chrominance in the image [14]. SLIC (Simple Linear Iterative Clustering) [15] which internally uses k-means clustering isolates the homogenous color regions in an image extending the opportunity to enlarge each region separately, by using a typical higher order interpolation technique restricting color mixing within neighboring regions. In order to handle the irregular shaped boundary of each of the isolated regions, a morphological dilation has been applied. The separately enlarged regions are then stitched together to form the final enlarged image.

The novel approach that is introduced through this study applies the particular algorithm separately on each of the 3 color channels of Y, Cb and Cr due to this specific color space being able to separate luminance and chrominance components. The isolation of luminance component can be advantageous where a higher resolution monochromatic sensor is available at the time of capturing the images. The effect of color bleeding will be significant in the chrominance components than in the luminance component resulting in only blurring out of the edges.

The paper is organized in the following manner. First similar works are discussed. Next the suggested method is explained in detail which is followed by the results which are discussed by depicting examples. Then, the evaluation method used is explained and finally the conclusion of the work is presented.

## II. RELATED WORK

The approach proposed through this paper focuses on using SLIC for segmentation and morphological analysis to perform interpolation at irregular shaped boundaries. This workflow greatly supports in mitigating the color bleeding effect explained above. This section discusses relevant and similar work carried out in this area.

### A. SLIC

SLIC is capable of clustering pixels on an image according to the color similarity and proximity on the image plane, forming nearly uniform superpixels [15]. Superpixels are a group of pixels with similar characteristics. The segmentation is performed on 5D space (i.e. vertical coordinate, horizontal coordinate, and 3 color channels) that internally uses k-means clustering with Euclidean distance for the distance measure used in the algorithm. This novel approach has been able to address issues present in other superpixel methods such as computational cost, poor segmentation quality, inconsistent size, shape etc. The distance measure introduced affects the compactness and regular shaping of the superpixel shape segments identified. We observed that SLIC is capable of separating homogeneous color regions. Therefore, SLIC is used in the proposed approach to isolate the homogeneous segments from each other.

### B. Morphological Processing

Morphological image processing performs non-linear operations based on a shape called as a structuring element or morphology of features on an image. Since the operation performed mainly relies on the pixel ordering rather than their color values, this is widely applied on binary and grayscale images [16].

The dimension of the structural element is usually odd to ease the process of performing functions on the origin when it is configured to be in the center. Erosion and dilation are two main operations performed using the concept of morphology. These are mainly used for identifying object shapes, separate touching objects and removing of unnecessary parts etc.

Dilation is the operation used in this proposed approach. This process makes an object grow in size. The width of growth depends on the shape and size of the structural element used. Dilation of grey-scale image $f$ by structural element B derives the result in equation (1),

$$\delta_B(f)_{(x,y)} = \max\{f(x-s, y-t) + B(s,t) | (x-s), (y-t) \in D_f, (s,t) \in D_B\} \quad (1)$$

where, (x,y) and (s,t) are respective coordinate sets and $D_f$ and $D_B$ are respective domains.

Image enhancement using morphological dilation mentioned in [17] is directly performed on the image by sliding a structural element over it. According to the position of the origin of the structural element on the image, a certain dilation is performed. They also emphasize the fact that the size of the structural element should be considered when performing the dilation as it affects the quality of the result.

### C. Interpolation techniques

1) *Nearest Neighbor Interpolation:* This is one of the simplest techniques that utilizes very limited execution time. Nearest Neighbour (NN) interpolation uses the existing values in the image to achieve the task and hence, refrains from producing any new color in the process. The pixel color values to be duplicated according to the enlargement factor are chosen by considering the distance between the new and the existing pixels. Although this does not introduce blurriness it produces an aliasing property to the image [18] making the result seem more artificial. Thus, the output image is of low quality and hinders further processing for various future applications.

2) *Bilinear Interpolation*: This method uses the distance-weighted average of the four nearest pixels to estimate the value of the target pixel. This is possible as the values represent a continuous function. For this, a linear interpolation is performed on each horizontal and vertical directions. According to Fig. 1, the color value of point P is affected by the four red Q points. Considering the relative distance between these 5 points, it is visible that $Q_{12}$ contributes more to decide the color of point P.

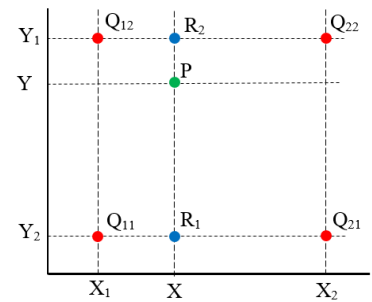

Fig. 1 Graph depicting bilinear interpolation

The end result from this approach is better than the NN, as the staircase effect introduced by NN is smoothened during the averaging process. The time complexity of this approach is moderate and thus is a better choice than the NN.

However, if the color of the pixel at point $Q_{12}$ (closest to P according to Euclidean distance) is composed of a contrasting color compared to the other 3 Q points, the resultant color at the interpolated point P would be outstanding and hence irrelevant. Thus, the presence of significantly odd colors as source points ($Q_{xy}$) would result in strange colors at interpolation which would lead to color bleeding around the blurred edges. The blurring effect contributes as a drawback to the image.

3) *Bicubic Interpolation*: Bicubic performs cubic interpolation in the two main axes of horizontal and vertical. Similar to bilinear interpolation, to estimate the value of the unknown pixel, the color value present on the nearest 16 pixels (4x4) are taken in to consideration in this method. As a result, a new color is assigned to the unknown pixel. Thus, comparatively this is a complex and advanced algorithm [18]. The intensity value assigned to point P(x,y) is derived as depicted in equation (2),

$$P(x,y) = \sum_{i=0}^{3} \sum_{j=0}^{3} a_{ij} \; x^i y^j \quad (2)$$

This approach has to determine 16 coefficients $a_{ij}$ from the 16 equations generated using this equation in order to determine the interpolated value.

A notable drawback depicted by this method is the increase of the acutance (the sharpness of the edge contrast) of images. The interpolation process results in enhancing the edges and thereby the result hinders visual perception.

### D. Deep learning approaches

Enhancing image quality by the use of Convolutional Neural Networks (CNNs) has been one of the most sought-after research areas in the recent past. One of the research that brought in valuable insights in to the community was carried out by Chao Dong, Chen Change Loy and Xiaoou Tang in [19]. Here the Super Resolution (SR) process makes use of a CNN to represent the functionality of the example-based SR on a method named Sparse-Coding as their internal algorithm in the image improvement process. Thus, an end-to-end mapping between the LR and HR images is learnt which contributes to make this unique from other existing external example-based methods. This methodology is also capable of performing SR on 3 channels of a color image simultaneously rather than focusing only on single-channel images.

Another notable approach [20] in this area is the image enhancement performed real-time for images captured using mobile devices. This model is trained on pairs of input/ output images performing computations on the bilateral grid and thereby approximating the affine color transformations.

Although deep learning approaches do address the SR issue, they showcase drawbacks such as high computational cost, domain dependency and the need for a large dataset to learn diverse features. Finding a balanced dataset for the targeted domain it-self is a challenge. When moving on to a separate domain the time-consuming process (training) needs to be repeated. Further the final results can be vulnerable if the dataset is biased.

### III. METHODOLOGY

This section explains the steps followed in the process of developing the proposed algorithm.

Fig. 2 displays the 4 main stages in the proposed algorithm in the process of creating a high-resolution quality enhanced image.

Novelty is achieved in this work through means on how SLIC and morphological dilation are performed together for the purpose of LR image enlargement. Generally, morphological dilation is performed on binary or grayscale images [16]. But in this work, the particular operation is performed parallel on each channel Y, Cb and Cr of the LR image to mitigate the color bleeding effect that may take place in the enhancement process.

- Preprocessing:

The algorithm is fed with a LR RGB image of the size 64x64 shown in Fig. 4. while the ground truth image is displayed in Fig. 3. The LR RGB is rescaled either using bicubic or bilinear interpolation in to a 256x256 image as seen in Fig. 5. Further, a padding is added to this image to address the issue of borders faced when sliding the window over the mask.

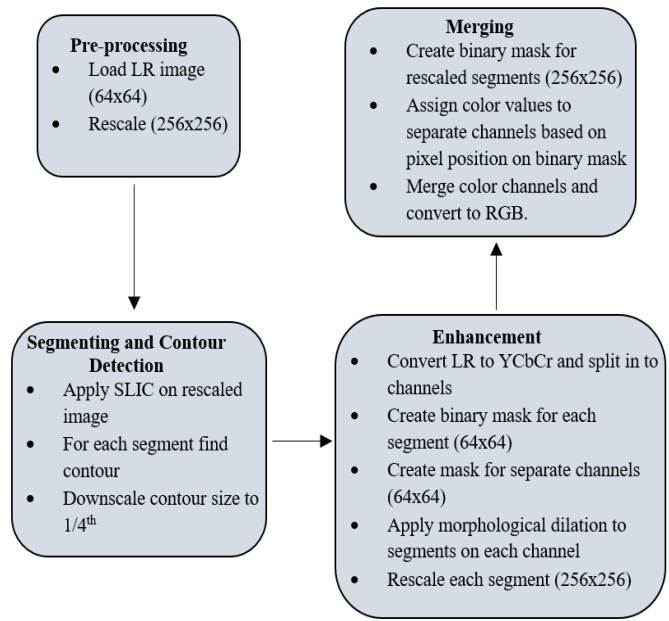

Fig. 2 Workflow of the proposed method

- Segmentation and contour detection:

As shown in Fig. 6 in the next phase SLIC is applied to the rescaled image to segment it according to homogeneous color regions. The number of color segments is configurable prior to execution. Then for each segment identified, it is thresholded according to the segment value and the relevant contours are calculated. The contour coordinates are scaled by the factor of 0.25 to match the size of the LR image.

- Enhancement:

In this stage a copy of the LR image is converted to YCbCr mode and the channels are split. Then a binary mask (64 x 64) "A" as in Fig. 7 is created for each segment identified earlier for the size of the downscaled contours. For the same set of contours, masks are created for each channel separately using the polygon fill method to retain the color values on each channel. Fig. 8 depicts the mask created for the Y-channel. In the same way channel-based masks are created for Cb and Cr channels as well.

A window of size 5x5 is used in this attempt. This window is slid across the mask A and if the origin (center) (x,y) of the window settles on the black region of this mask, the maximum color value read on the 5x5 window on the corresponding mask (B) at the same position of (x,y) is dilated. Else the window continues moving to the next pixel in line. The reason to take this approach is to avoid color bleeding (for each channel) within the segments in the B and the black background when a direct rescale to 256x256 is performed. Fig. 9 shows a dilation for the 1st segment of the Y-channel.

After the dilation, each of the segments are rescaled to 256x256 as seen in Fig. 10.

- Merge:

As the final step a binary mask is created for the rescaled 256x256 output from the previous stage for the size of the original contours. For each pixel on this mask if the pixel color is white on the mask, the corresponding values on the separate channeled masks will be collected to form a new image. Finally, all the channels are merged (stitched) and converted to RGB.

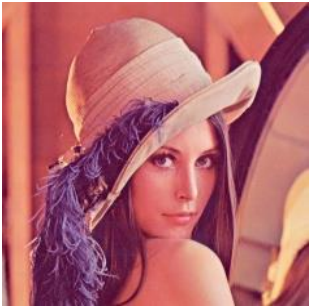
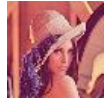

Fig. 3 Ground truth image          Fig. 4 Down-scaled 64x64 image

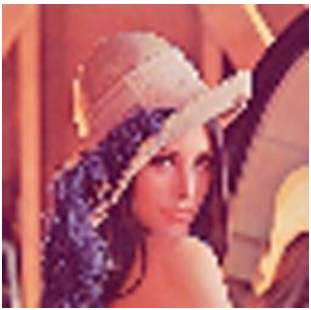
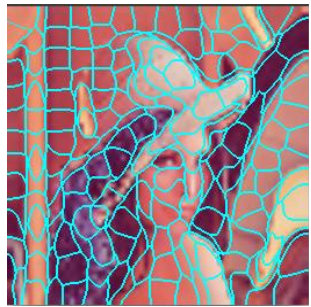

Fig. 5 Bicubic interpolated image (256x256)          Fig. 6 Segments after applying SLIC

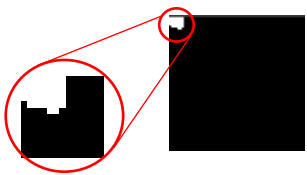
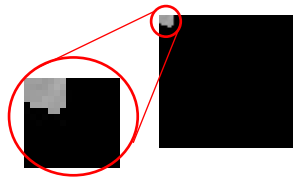

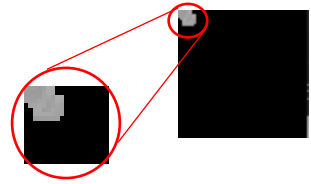
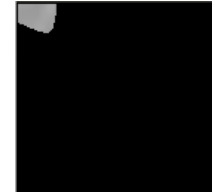

Fig. 7 Binary mask for segment1 on 64x64 image          Fig. 8 Corresponding mask for Y channel

Fig. 9 Morphology dilated Y channel mask (64x64)          Fig. 10 Up-scaled to 256x256

IV. RESULTS

The results obtained from this study are discussed in detail below.

*A. Dataset*

The dataset used for the evaluation process of the algorithm includes the standard Set 5 and Set 14 [21] image datasets and 10 random images chosen based on the color and shape complexity. The random images are available at [22].

*B. Results*

The results show that the jagged nature of edges that is generally introduced when using traditional interpolation techniques such as bicubic or bilinear interpolation is not present when following the proposed methodology. The edgy nature of lines is mitigated and an eye-pleasing and a quality output is received as a result.

Fig. 11 and Fig. 12 display the ground truth image (a) against the traditional interpolated image (b) and the output received from the proposed method (c). The separately enlarged parts visibly illustrate how the jagged edges are attended to by the method suggested through this work.

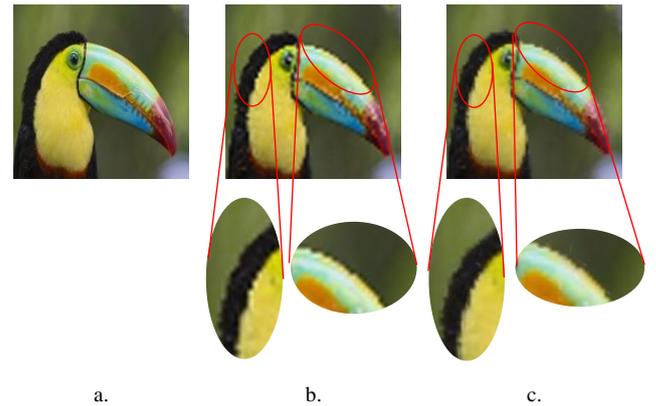

a.          b.          c.

Fig. 11 Example output comparing the jagged edges.
**a**. the ground-truth, **b**. bicubic interpolation, **c**. SLIC based bicubic enlargement

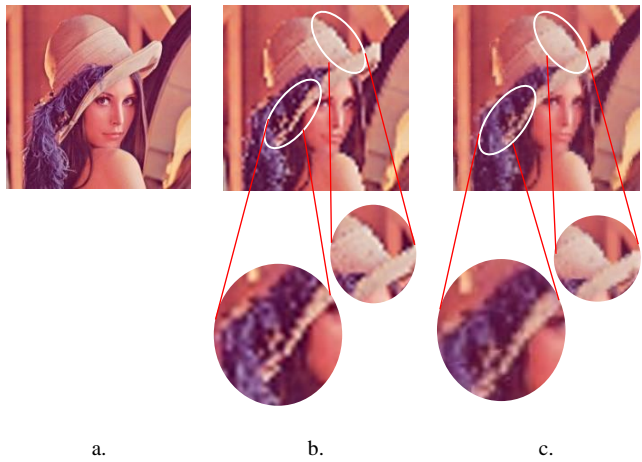

a.　　　　　　　b.　　　　　　　c.

Fig. 12 Example output comparing the jagged edges of Lena image. **a**. the ground-truth, **b**. bicubic interpolation, **c**. SLIC based bicubic enlargement

Further, it is visible that the separately processed channels after merging has not introduced color bleeding or color displacement in to the result. The segments have been enriched separately without disturbing the colors around the edges. Hence each segment displays clear cut boundaries and solid color spread across the segments.

The PSNR values shown in the tables below obtained in contrast to the traditional bicubic and bilinear methodologies clearly confirms that the quality of the result obtained from the proposed method is better. Table 1 depicts results obtained from the random images chosen and Table 2 shows the average values taken from the Set 5 and Set 14 dataset.

TABLE I
COMPARING PSNR (DB) VALUES OF DIFFERENT APPROACHES

|  | Bicubic | SLIC based bicubic | Bilinear | SLIC based bilinear |
|---|---|---|---|---|
| Image 1 | 23.20 | 27.08 | 23.72 | 27.09 |
| Image 2 | 20.64 | 24.01 | 21.31 | 24.00 |
| Image 3 | 19.92 | 24.73 | 20.43 | 24.77 |
| Image 4 | 22.28 | 26.21 | 22.88 | 26.21 |
| Image 5 | 20.59 | 25.26 | 21.22 | 25.25 |
| Image 6 | 19.69 | 22.71 | 20.23 | 22.64 |
| Image 7 | 19.99 | 24.50 | 20.67 | 24.50 |
| Image 8 | 17.00 | 19.67 | 17.63 | 19.67 |
| Image 9 | 16.28 | 20.35 | 16.90 | 20.37 |
| Image10 | 23.15 | 26.70 | 23.52 | 26.70 |
| Average | 20.274 | **24.122** | 20.851 | 24.12 |

TABLE 2
COMPARING AVERAGE PSNR (DB) VALUES OF SET 5 AND SET 14

|  | Bicubic | SLIC based bicubic | Bilinear | SLIC based bilinear |
|---|---|---|---|---|
| Set 5 | 20.48 | **25.00** | 21.04 | 24.98 |
| Set 14 | 18.43 | **21.69** | 19.06 | 21.67 |

According to the average values obtained, the performance of SLIC based bicubic and SLIC based bilinear are quite similar. As the proposed method applies the interpolations only over the homogeneous color regions, this result also reveals that both bilinear and bicubic method perform similarly in the absence of strong edges and the real advantage of the bicubic interpolation can be seen only with the presence of strong edges or regions with contrastive colors. However, the result shows a slightly better performance when bicubic interpolation is combined with SLIC in the proposed method.

This performance is achieved by reusing the existing basic and advanced methods in image enhancement. However, the way how these techniques are chained and utilized brings in the novelty in to the approach and hence, affects the quality of the output significantly. The traditional bilinear and bicubic interpolation assists in image interpolation using comparatively complex calculations but yielding far better outputs compared to nearest-neighbor. But, both the methods tend to introduce color bleeding in the process of enhancement. The proposed solution in contrast to the above, successfully handles the edges by preventing color bleeding. This is achieved by handling different homogenous regions separately by using SLIC and enlarging them in an isolated manner using morphology and finally stitching them together. Furthermore, the procedures performed on each separate color channel Y, Cb and Cr has contributed to produce more refined and clear results. This is due restricting color bleeding in each of the channels individually.

## V. EVALUATION

For this work a quantitative analysis was performed to gain an understanding about the quality of the result obtained after applying the proposed method.

High resolution images were re-scaled to the LR size of 64x64 and then either using bilinear or bicubic interpolation they were re-scaled to 256x256. Then the SLIC based image enlargement algorithm was applied on them and the corresponding results were used to calculate the PSNR values. The down-scaled original HR image was compared with the interpolated and SLIC based enhanced images separately during the analysis. The proposed algorithm was run on 2 standard datasets and 10 test images to evaluate the performance of it on different color distributions and shapes.

Apart from the numerical comparison, the visual comparison of the results also reveals less aliased and non-color-bleeding edges with preserving the acutance. The difference between the SLIC based bicubic and SLIC based bilinear is also visible on close inspection.

## VI. CONCLUSION

This paper has focused to discuss on the need for high resolution images and how to reconstruct them from low resolution images available. It has also been able to bring out a method to perform digital image enlargement which carefully tackles the color bleeding issue present in other baseline methods. The proposed solution uses SLIC and morphological dilation to achieve image enhancement that is capable of producing quality and acceptable images given a low-resolution image. This algorithm is applied to separate channels of luminance and chrominance and the processed channels are combined at the final step. The quantitative analysis depicts that the results for the proposed method outperforms the existing ones yielding better quality images.

VII. FURTHER WORK

It is visible through this work that, the dimension of the structural element can contribute significantly in the reconstruction process. Along with the dimension of the structural element the number of SLIC segments configured can also be changed to observe whether a performance improvement can be gained in the resultant image. Further, other interpolation techniques other than the two basic baseline methods used in this approach can be used and tested for the performance.